\documentclass{llncs}
\usepackage{graphicx,url}
\usepackage{xcolor,wrapfig,algorithmic}
\usepackage[scriptsize,tight]{subfigure}
\usepackage{tabularx}
	\newcolumntype{C}[1]{>{\centering\arraybackslash}p{#1}}
	\newcolumntype{R}[1]{>{\raggedleft\arraybackslash}p{#1}}
	\newcolumntype{L}[1]{>{\raggedright\arraybackslash}p{#1}}
\usepackage{booktabs}	
\usepackage[ruled,linesnumbered]{algorithm2e}
\usepackage{amssymb}
\usepackage{xspace}
\newcommand{\B}{\mbox{$\mathbb{B}$}}
\newcommand{\etal}{\emph{et al.}\xspace}

\begin{document}

\date{}

\title{Evolution of Convolutional Highway Networks}

\author{Oliver Kramer}

\institute{Computational Intelligence Group\\Department of Computer Science\\University of Oldenburg, Germany}

\maketitle
\thispagestyle{empty}

\begin{abstract}
Convolutional highways are deep networks based on multiple stacked convolutional layers for feature preprocessing. We introduce an evolutionary algorithm (EA) for optimization of the structure and hyperparameters of convolutional highways and demonstrate the potential of this optimization setting on the well-known MNIST data set. The (1+1)-EA employs Rechenberg's mutation rate control and a niching mechanism to overcome local optima adapts the optimization approach. An experimental study shows that the EA is capable of improving the state-of-the-art network contribution and of evolving highway networks from scratch.
\end{abstract}

\section{Introduction}

Convolutional networks are extraordinary successful in many domains, e.g., in image recognition. Convolutional highways~\cite{highways} allow training of convolutional networks with large number of layers and have been introduced as counterparts of long short term memory (LSTM)~\cite{LSTM} networks. Each convolutional highway layer employs two gates for the flow of information, i.e. for convolution and for passing information. For novel applications, the optimal network structure and hyperparamterization is often unknown. The application of evolutionary heuristics for finding optimal or near-optimal networks has a great potential~\cite{bigstep}. Examples for the use of EAs are the optimization of network structures, of layers, their composition (modules), and of hyperparameters. 

At the end of the nineties, neuroevolution was a successful research direction, but mostly concentrated on the evolution of connections between neurons and the number of neurons in layers of multilayer perceptrons (MLPs). The latter are today mostly used as last layers in convolutional networks known as dense or fully connected ones. The objective of this paper is to show that a (1+1)-EA with Rechenberg mutation rate control and niching does an excellent job in network optimization, e.g., for evolving deep networks from scratch in unknown domains.

This paper is structured as follows. In Section~\ref{sec:high} we introduce convolutional highways. Related work on optimization of deep learning networks is discussed in Section~\ref{sec:related}. The optimizing EA is introduced in Section~\ref{sec:evo}, and experimentally analyzed on the MNIST data set in Section~\ref{sec:exp}. Results are summarized in Section~\ref{sec:cons}, where also the role of EAs in deep learning is discussed.

\section{Convolutional Highways}
\label{sec:high}

Convolutional neural networks have been introduced by LeCun \etal~\cite{lecun}. They are based on convolutional layers, which consist of three parts. The first part applies filter/kernels (small weight matrices) that are convolved with the input $\mathbf{x}$ by matrix multiplication. The part of the input that is convolved with the kernel is called receptive field. The result of the convolution process is written into the activation map. The activation map is subject to a pooling process, which reduces the dimensionality, for example by employing the maximum of a rectangular (max pooling). This process if followed by an activation layer applying a non-linear function. For example, the activation function ReLu (rectified linear unit) turns all negative numbers to 0. A classic convolutional network consists of three such convolutional layers followed by two or more fully-connected dense layers.

Convolutional highways are based on two ideas. First, they stack multiple convolutional layers for feature preprocessing. Second, each convolutional highway layer employs two gates for the flow of information. A shared convolutional gate employs the usual convolutional layer, which is combined with a weight matrix $\mathbf{W}_C$. A transform gate controls the amount of information that is passed through the convolutional layer employing a transform matrix $\mathbf{W}_T$. The inverse, i.e., $\mathbf{1} - T(\mathbf{x},\mathbf{W}_T)$ defines the amount of information of input $\mathbf{x}$ that is passed through the layer. Hence, one convolutional highway layer outputs (leaving out the bias for the sake of readability)
\begin{equation}
\mathbf{y} = H(\mathbf{x},\mathbf{W}_C) \cdot T(\mathbf{x},\mathbf{W}_T) + \mathbf{x} \cdot (\mathbf{1}-T(\mathbf{x},\mathbf{W}_T)).
\end{equation}
The highway network is composed of multiple modules each consisting of $k$ succeeding convolutional layers with decreasing kernel size followed by max pooling and normalization. The convolutional highway layers are followed by dense layers and a final softmax layer. The network structure is illustrated in Figure~\ref{fig:highway}.

The EA adapts the number of convolutional highway layers within each module, the number of modules and hyperparameters like kernel sizes and activation function types. Details are presented in Section~\ref{sec:neuroevo}.

\section{Related Work}
\label{sec:related}

The line of research on neuroevolution began in the nineties with many interesting approaches, of which most concentrated on the number of neurons and the structure of MLPs. One of the most famous contributions in this line of research is NEAT~\cite{neat}, which is able to evolve MLPs employing techniques like augmenting topologies and niching. Its successor HyperNEAT~\cite{hyperneat} is able to evolve networks, but does not achieve state-of-the-art performance. 
Compositional pattern-producing networks (CPPN)~\cite{cppn} assume the general network structure is predefined, but its components are independent of each other. Fernando \etal~\cite{gecco2016} extend CPPN for autoencoders by a Lamarckian approach that inherits the learned weights. Ilya Loshchilov and Hutter~\cite{cma} employ the CMA-ES to evolve the hyperparameters of convolutional networks optimizing dropout and learning rates, batch sizes, numbers of filters, and numbers of units in dense layers. Suganuma \etal~\cite{gecco2017} propose a genetic programming (GP) approach for designing convolutional networks achieving competitive results to state-of-the-art convolutional networks. Recently, Real \etal (Google)~\cite{googleEA} invested exhaustive evolutionary search to evolve convolutional networks for image classification on CIFAR. Also LSTM cells have been subject to evolutionary architecture search by J{\'{o}}zefowicz \etal (Google)~\cite{evoLSTM}.

Recent related approaches by Bello \etal (Google)~\cite{googlebrain} and Baker (MIT) \etal~\cite{MIT} employ reinforcement learning for evolving deep convolutional networks. Machine learning pipelines can be evolved with EAs for example with the tree-based pipeline optimization tool (TPOT) by Olson \etal~\cite{TPOT} or for kernel PCA pipelines~\cite{ki2017}, for which an integer-based representation has been used.

\section{Evolutionary Approach}
\label{sec:evo}

Evolutionary algorithms are powerful tools for blackbox optimization problems with local optima. While finding countless successful applications, from numerical to structural optimization, EAs are grounded on a solid theoretical basis, see e.g.~\cite{theorybook}.

\subsection{(1+1)-EA}

For network evolution we employ a (1+1)-EA that generates a new child $\mathbf{z}' \in \B^N$ in binary representation with bit string length $N$ based on a single parent $\mathbf{z}$ in one generation with bit flip mutation. If the fitness of the child is better than the fitness of its parent, it replaces its parent. The process is repeated until a termination condition is met. As we represent the phenotype of the convolutional highway as bit string, the EA employs bit flip mutation with probability $\sigma$. The EA uses mutation rate control and niching, see Algorithm~\ref{alg:ea}.

\begin{figure}[h!]
\center
\begin{minipage}{1.\linewidth}
\begin{algorithm}[H]
\caption{(1+1)-ES} \label{alg:ea} 
\center
\begin{algorithmic}[1]		
	\STATE intialize~$\mathbf{z} \in \B^n$ randomly
	\REPEAT
	\STATE \textbf{if} niching$\_$mode = \textbf{true} for $\kappa$ gen. \textbf{and} $f(\mathbf{z}) < f(\mathbf{z}_n)$
	\STATE \hspace{6mm} replace~$\mathbf{z}$ with~$\mathbf{z}_n$, niching$\_$mode = \textbf{false}
	\STATE mutate~$\mathbf{z} \rightarrow \mathbf{z}'$ with bit flip
	\STATE adapt $\sigma$ with Rechenberg
	\STATE replace~$\mathbf{z}$ with~$\mathbf{z}'$ \textbf{if}~$f(\mathbf{z}) \geq f(\mathbf{z}')$
	\STATE \textbf{else} with probability $\eta$ $\mathbf{z}_n = \mathbf{z}$, replace~$\mathbf{z}$ with~$\mathbf{z}'$,
	\STATE \hspace{33mm} count gen. with $\kappa$, niching$\_$mode = \textbf{true}
	\UNTIL termination condition
			\end{algorithmic}
			\end{algorithm}
	\end{minipage}
\end{figure}

\subsection{Mutation rate control}

To adapt the mutation rate during the optimization process, the Rechenberg rule is applied. Rechenberg's rule adapts the mutation rate according to the success rate of the EA. In case of high success rates with $g/G\geq1/5$ and number $g$ of successful generations in a window $G$, the mutation rate $\sigma$ is increased by multiplication $\sigma' = \sigma \cdot \tau$ with $\tau > 1$. Otherwise for $g/G < 1/5$, $\sigma$ is decreased with $\sigma' = \sigma / \tau$. The rule allows making larger steps in case of sequent success and smaller steps in case of stagnation.

\subsection{Niching for (1+1)-EA}

To overcome local optima, we introduce a niching approach for the (1+1)-EA, which suffers more from multi-modality than population-based EAs. If a child is worse than its parent, it is not rejected with probability $\eta$, but optimized for $\kappa$ generations. The fitness of the last child of this optimization branch replaces the last parent in the main optimization branch, if it employs a better fitness. Otherwise, the original parent is the basis for the subsequent optimization run.

\begin{figure}[h]
\begin{center}
\includegraphics[scale=0.45]{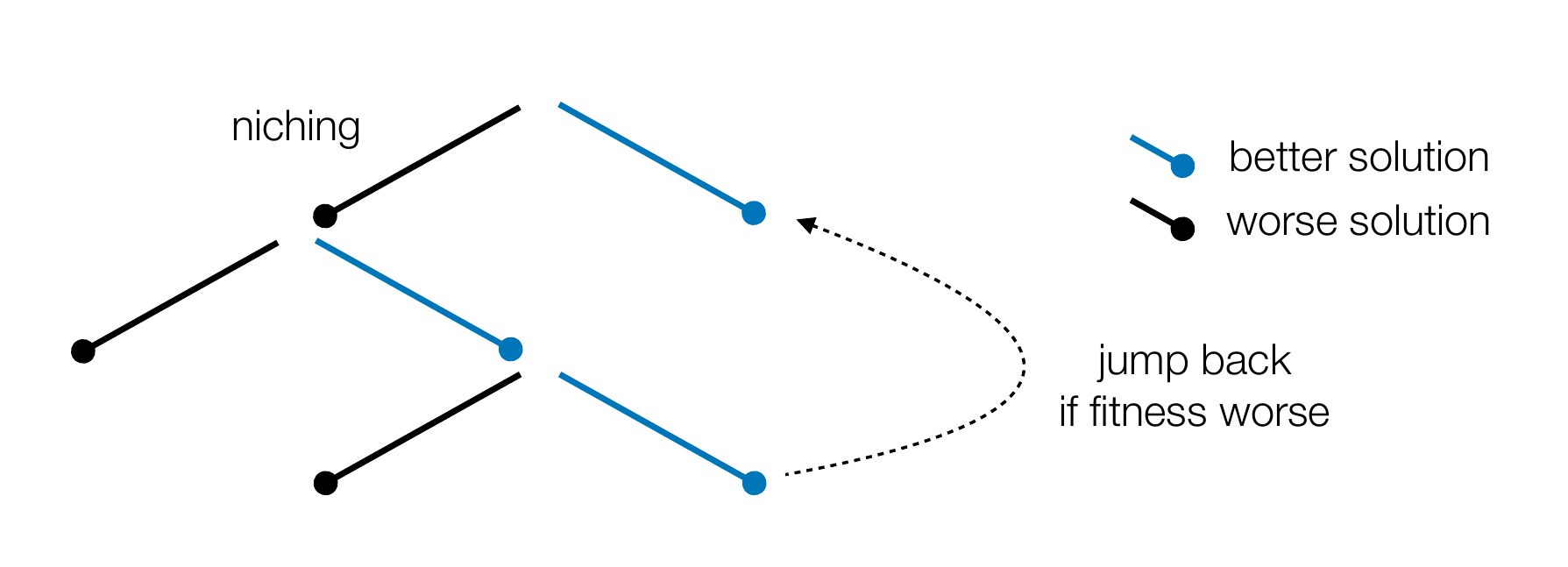}
\caption{\label{fig:niching}The niching process allows the (1+1)-EA to follow a worse solution with probability $\eta$ for $\kappa$ generations.} 
\end{center}
\end{figure}

\subsection{Network evolution}
\label{sec:neuroevo}

\begin{figure}[h]
\begin{center}
\includegraphics[scale=0.4]{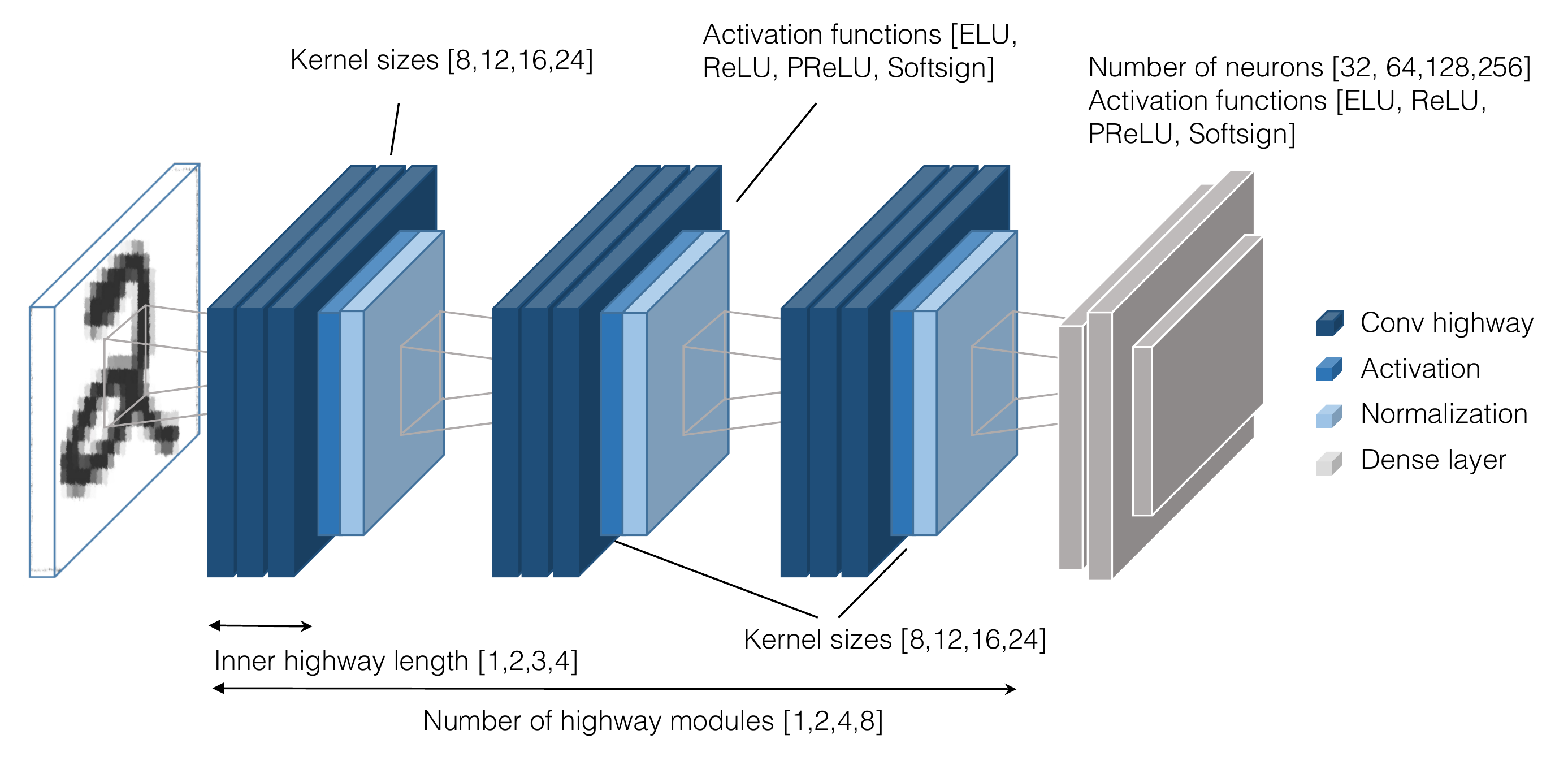}
\caption{\label{fig:highway}The evolutionary convolutional highway network allows an adaptation of the number of highway modules, their inner module structures, and the network hyperparameters.} 
\end{center}
\end{figure}

The EA optimizes the convolutional highway structure and its hyperparameters, see Figure~\ref{fig:highway}. Concerning the network structure the EA adapts the number of highway modules ($1,2,4,8$). Within each module it further adapts the number of convolutional 2d layers ($1,2,4,8$). The numbers of neurons for the two dense layers are also optimized (from set $32,64,128,256$). Further, the EA adapts hyperparameters like the kernel size of all highways ($8,12,16,24$), the kernel size of the max pooling layers ($1,2,3,4$), and the activation function types of all highways and of the dense layers (ELU\footnote{$x$ for $x>0$ and $\alpha(e^{x}-1)$ for $x \leq 0$}, ReLU\footnote{$\max(x,0)$}, PReLU\footnote{parametric ReLU}, Softsign\footnote{$x/(|x|+1)$}). Last, the EA evolves the learning rate of the network apply \textit{Adam} as gradient descent optimizer.

The convolutional highway network is represented as bit string. It is translated to a phenotype by piecewise mappings to integers, which are used as indices for lists containing parameterizations. We use TFlearn to build a TensorFlow model that is executed for each fitness function evaluation. The network is trained using cross-validation. The final classifier is evaluated on the independent test set, on which the categorical cross-entropy is computed as fitness value. In total, 20 bits are used to represent a convolutional highway network resulting in an overall solution space size of over one million networks.

\section{Experimental Study}
\label{sec:exp}

The experimental study concentrates on the evolution of convolutional highways from scratch, e.g., based on a completely random solution. Our experiments are based on the well-known image recognition data set MNIST with training set size 55,000 and test set size 10,000. The (1+1)-EA runs for 30 generations. The (1+1)-EA with Rechenberg employs the settings $G=10$ and $\tau = 0.5$, the niching mechanism uses $\eta = 0.1$ and $\kappa=10$. These are settings that turned out to work well on pre-experiments. Each network is trained for 5 epochs. Table~\ref{tab:param} summarizes the parameter settings of our study.

\begin{table}[htb]
\caption{\label{tab:param}Parameter settings of EA and convolutional highway network}
\center
\begin{tabular}{| l | l | l | l | l | l |l |l |l |l |l |l |l |l |l |l |}

\hline
\multicolumn{2}{|l}{EA}& \multicolumn{2}{|l|}{highway network}\\
\hline
parameter & value & parameter & value \\
\hline
init. mutation rate $\sigma$ 		& 1/N 	& type					& conv. highway \\
Rechenberg G		 				& 10 	& epochs				& 5 \\
Rechenberg $\tau$ 				& 0.5 	& learning rate			& evolved \\
niching $\eta$ 						& 0.1 	& gradient descent 		& \textit{Adam}\\
niching $\kappa$					& 10 	& error / loss 			& categorial cross-entropy\\
generations							& 30 & init. 				& random \\
\hline
\end{tabular}
\end{table}

\begin{figure}[htb]
\begin{center}
\subfigure[standard\label{subfig:net1}]{\includegraphics[scale=0.36]{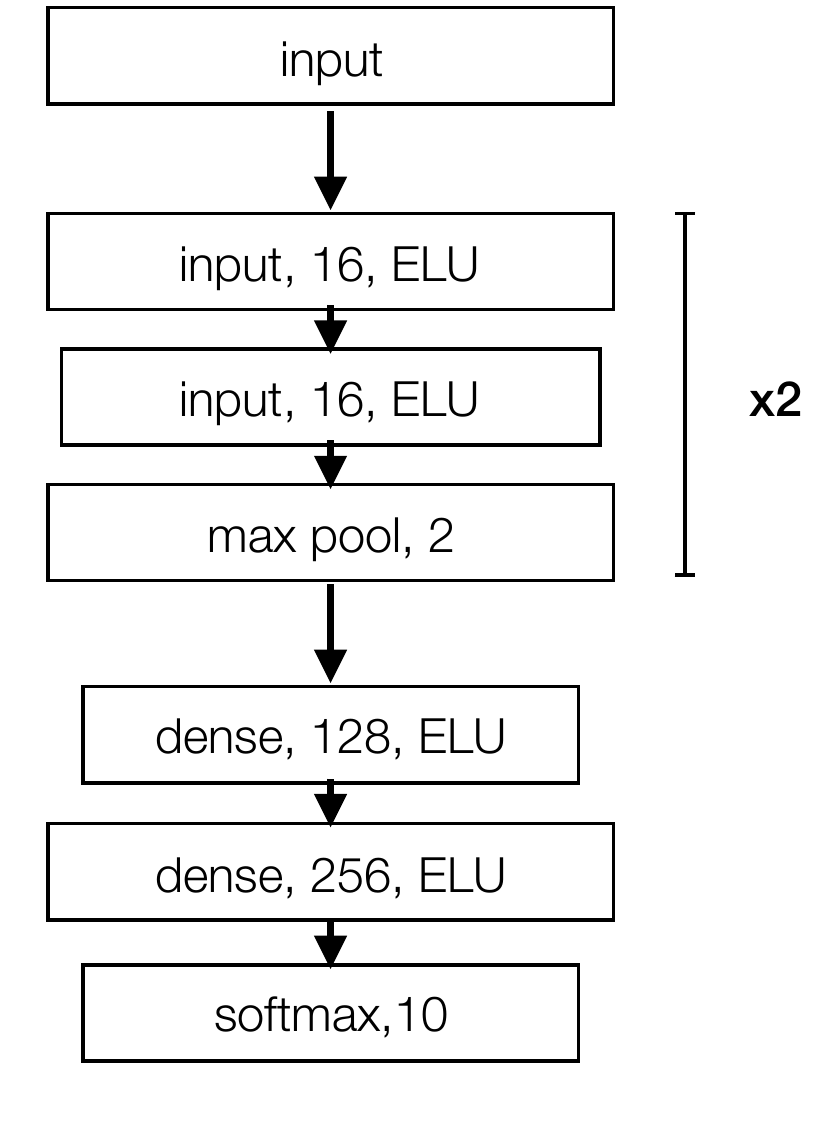}}
\subfigure[best MNIST\label{subfig:net2}]{\includegraphics[scale=0.36]{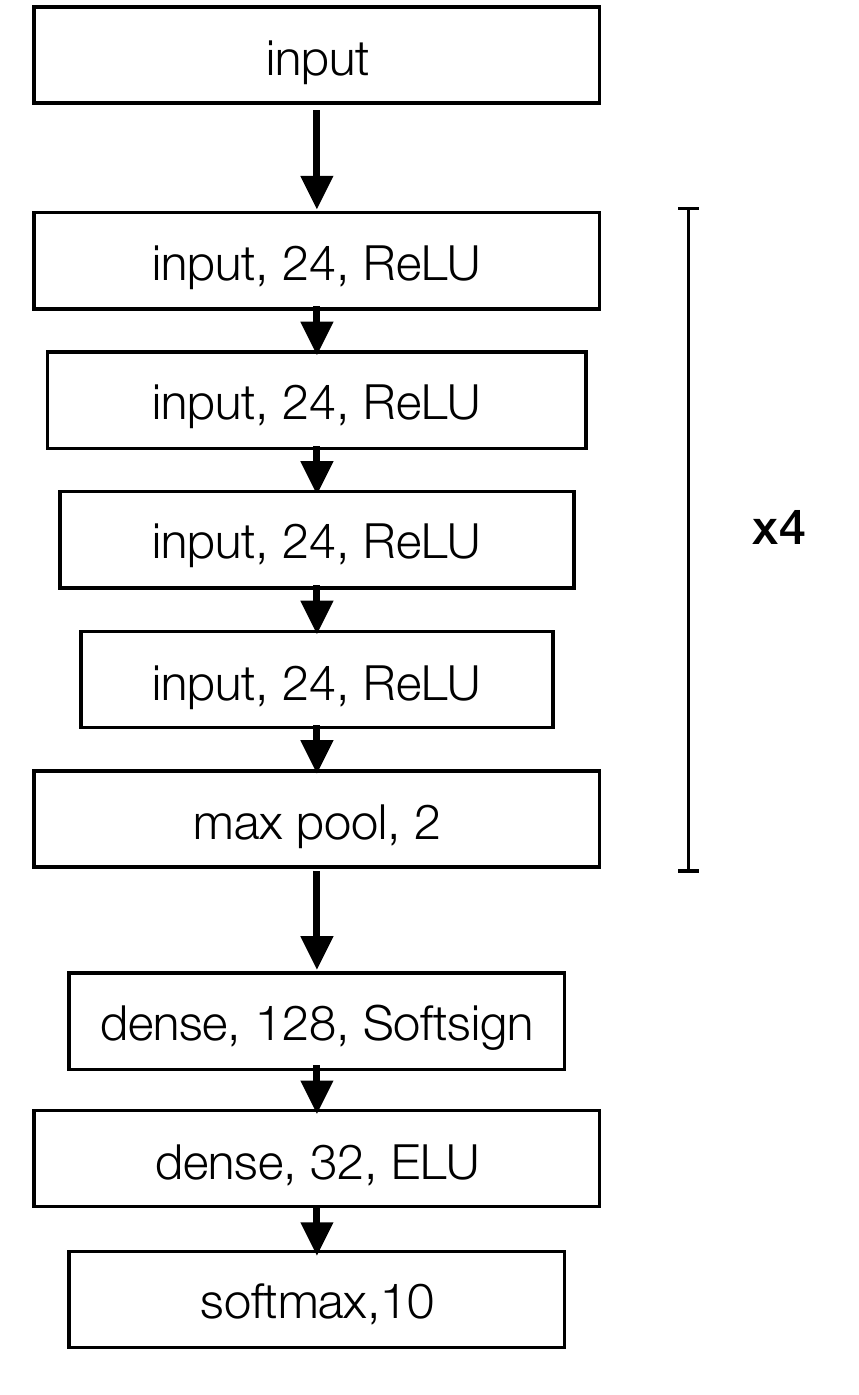}}
\caption{\label{fig:nets}The comparison between the standard convolutional highway network and the best highway variant evolved.} 
\end{center}
\end{figure}

Table~\ref{tab:exp} shows the experimental study of the evolved convolutional network. For comparison, the error of the network with standard specifications (like employed in TFlearn) is shown, see Figure \ref{subfig:net1}, and the median fitness of the first random initial networks. All runs are repeated 10 times. 
\begin{table*}[htb]
\caption{\label{tab:exp}Experimental study of convolutional highway accuracy on MNIST of the (1+1)-EA in the simple version, with Rechenberg mutation rate control, and with the niching mechanism.}
\center
\begin{tabular}{| l | l | l | l | l | l |l |}
\hline
(1+1)-EA & standard & median init. & min & mean & std & max \\
\hline
simple 		& 	0.977 	&	0.979	&	0.972 	& 0.983			& 0.007		& 0.989		\\
Rechenberg 	& 	0.977 	&	0.917	& 0.941		& 0.970			& 0.020 	& 0.986		\\
niching		& 	0.977 	&	0.973 	& 0.986 	& 0.989			& 0.001		& 0.991	\\
\hline
\end{tabular}
\end{table*}
The errors of the final best network and the mean of the best networks of all runs are presented. The results show that the (1+1)-EA is able to evolve the convolutional highway network from scratch. 
The random initial networks achieve significantly worse results than the networks with recommended standard setting, but the EA is able to improve the networks in all cases. The (1+1)-EA with Rechenberg performs worse than the standard (1+1)-EA. But the variant with Rechenberg and niching beats both other types in best, mean, and worst results, while achieving stable performance with a small standard deviation.

\begin{figure}[h!]
\begin{center}
\subfigure[simple\label{subfig:runs1}]{\includegraphics[scale=0.19]{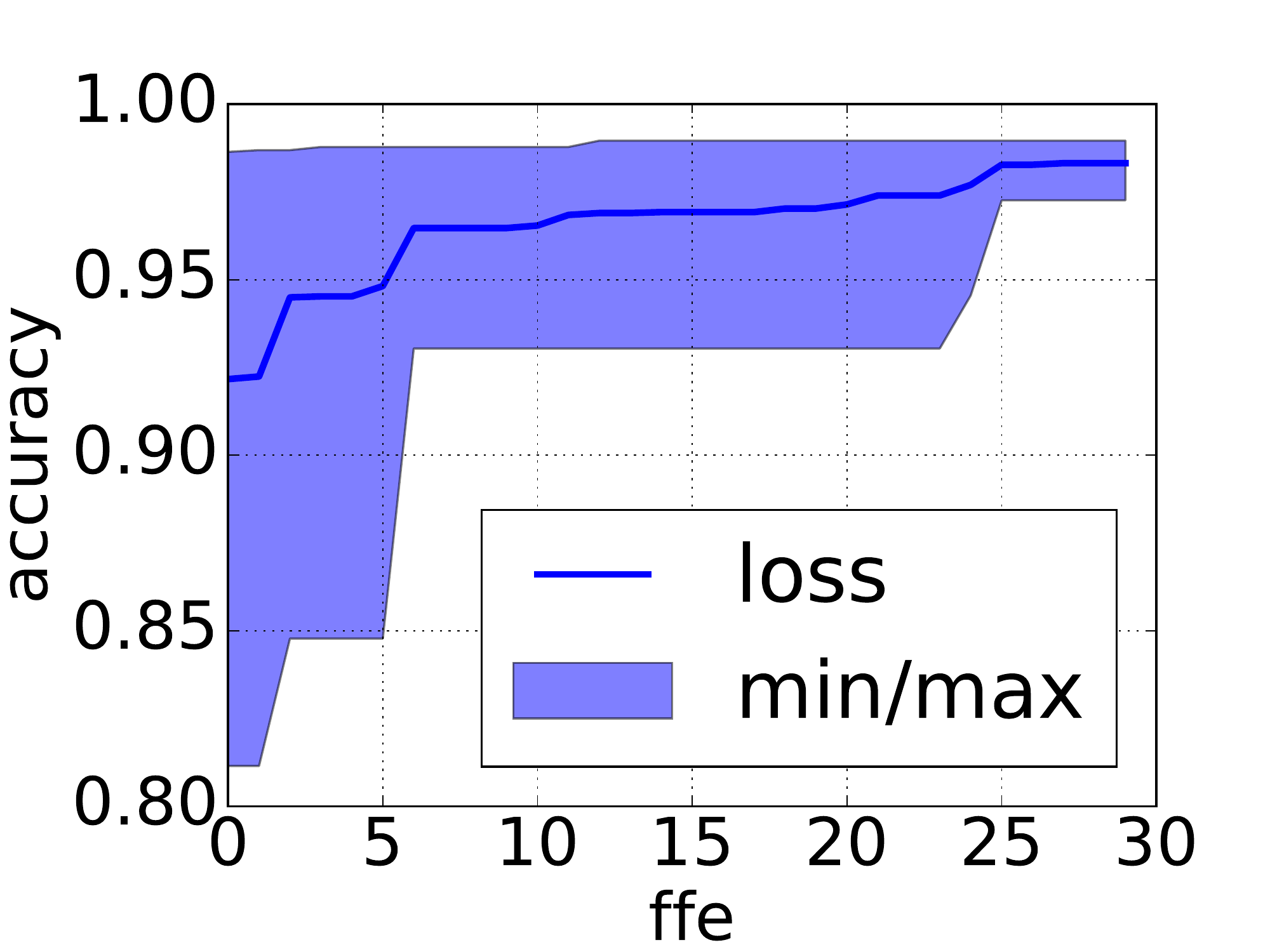}}
\subfigure[Rechenberg\label{subfig:runs2}]{\includegraphics[scale=0.19]{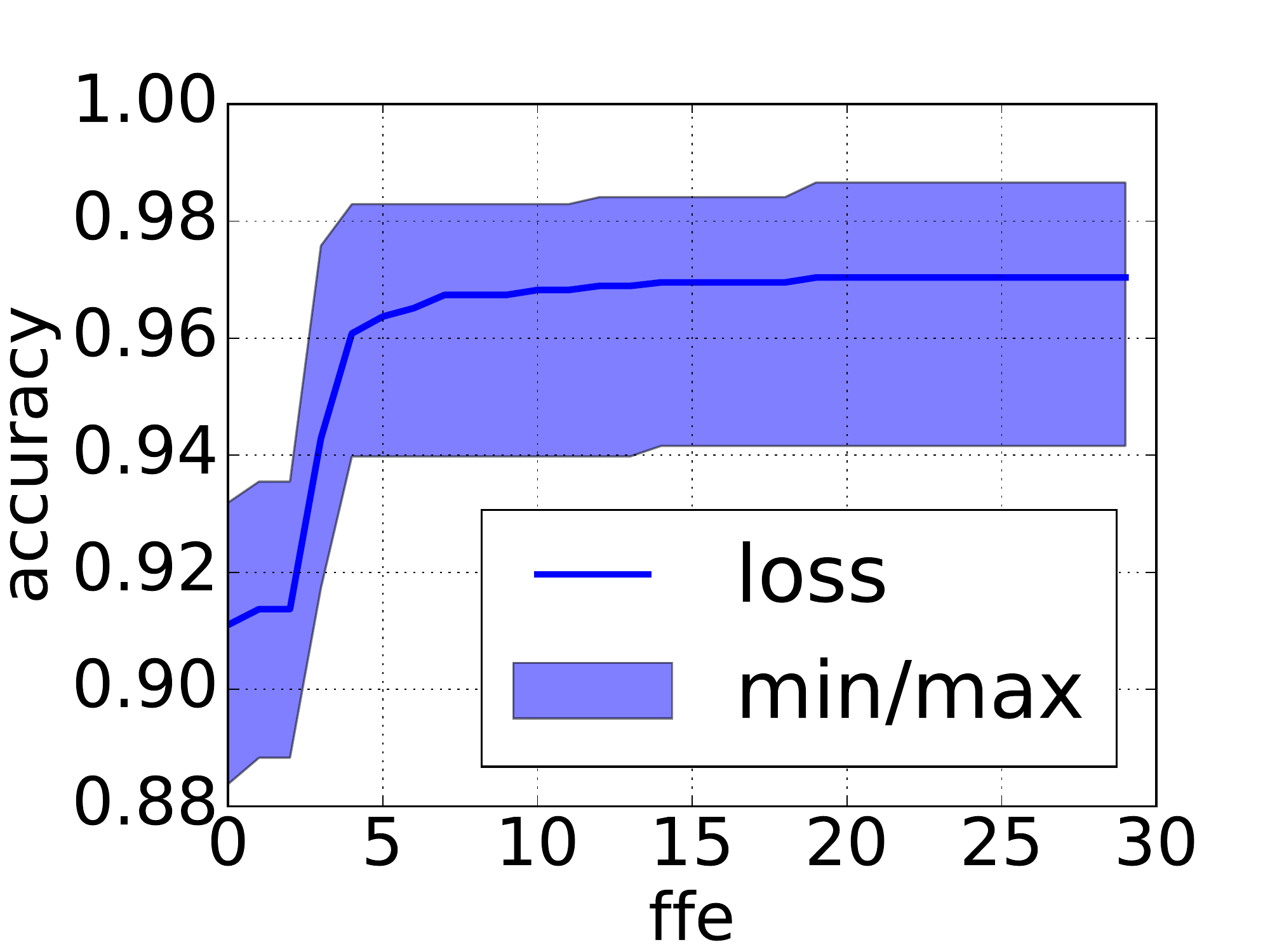}}
\subfigure[niching\label{subfig:runs3}]{\includegraphics[scale=0.19]{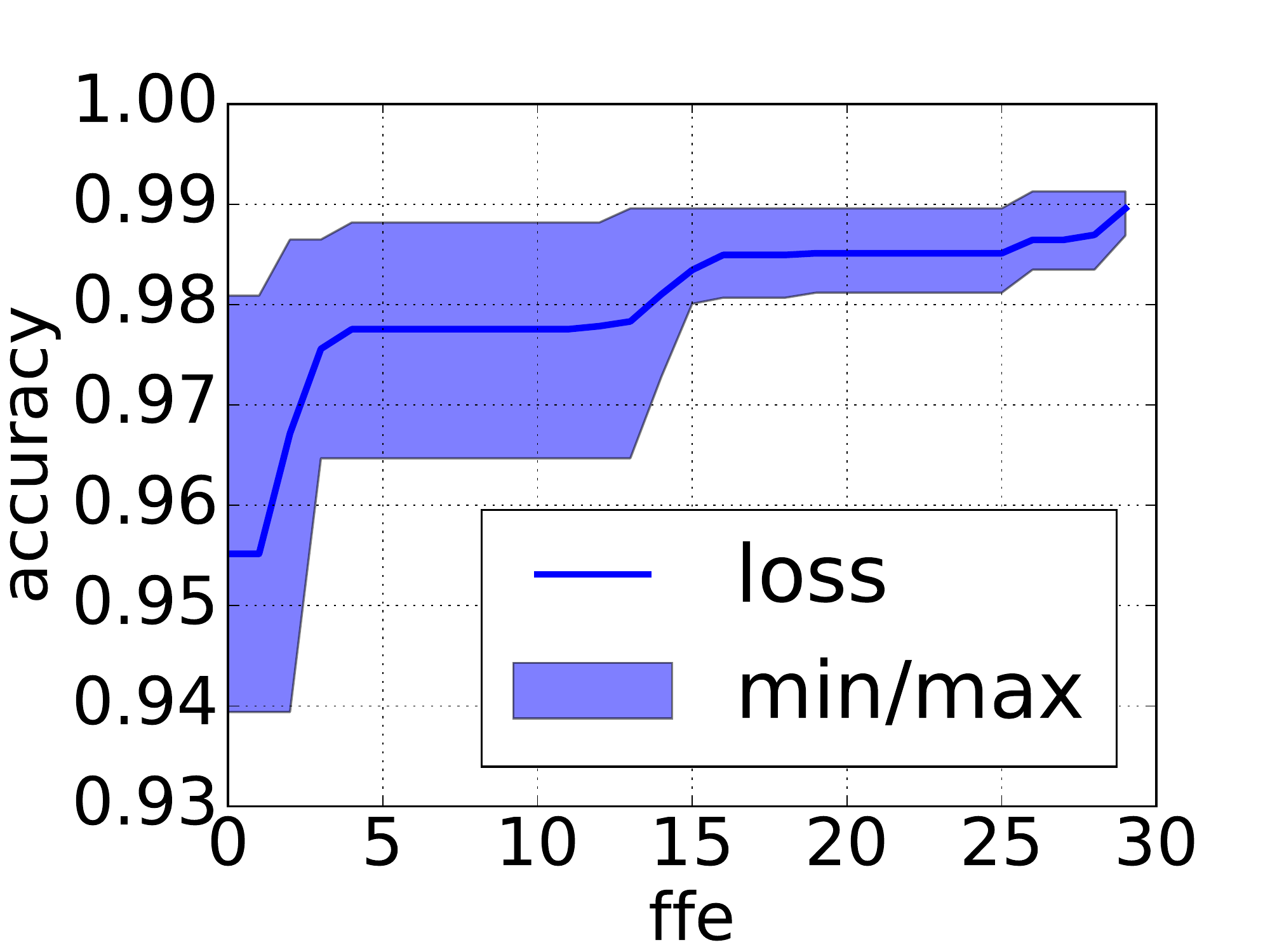}}
\caption{\label{fig:runs}Fitness developments of the (1+1)-EA variants evolving convolutional highways for 30 generations.} 
\end{center}
\end{figure}

Figure~\ref{fig:runs} shows the fitness development of the (1+1)-EA variants optimizing the convolutional highways on MNIST. All runs show a stable convergence towards values over 0.94 accuracy level. Also the worst initial nets can be adapted by the EA to achieve competitive performance.

Figure~\ref{fig:nets} compares the best evolved net to the standard network. The best net has been optimized with the (1+1)-EA, Rechenberg's mutation rate control, and niching. The comparison shows that the EA has evolved significantly different network structures.

\section{Conclusions}
\label{sec:cons}

This work demonstrates that a comparatively simple EA with mutation rate control and a niching mechanism is able to evolve convolutional highways from scratch, i.e., from random initializations.
The evolved network structures are significantly different from standard networks in deep learning frameworks. The results allow the conclusion that EAs are powerful techniques for optimization of highway network structures and hyperparameters. Mutation rate control and niching are helpful to support the optimization process. 

The application of EAs to network learning has numerous advantages. EAs do not require gradients, prior assumptions regarding problem knowledge, or human expertise. Further, EAs are embarassingly parallelizable, allow unexpected results, and can easily optimize two or more conflicting objectives at a time. As future work we plan to apply the evolutionary convolutional highways to domains like video and speech recognition. Further, we plan to extend the experiments to large-scale data sets.

\appendix


\bibliographystyle{abbrv}

\end{document}